\documentclass[a4paper,conference]{IEEEtran}

\ifCLASSINFOpdf
\else
\fi

\usepackage{latexsym}
\usepackage{multirow}
\usepackage{algorithm} 
\usepackage{algpseudocode} 
\usepackage{multicol}
\usepackage{comment}
\usepackage{hyperref}
\usepackage{amsmath}
\usepackage{graphicx}
\usepackage{subcaption}
\usepackage{url}

\begin{document}
%
\title{Sequential Domain Adaptation through \\ Elastic Weight Consolidation for Sentiment Analysis}

\author{\IEEEauthorblockN{Avinash Madasu}
\IEEEauthorblockA{Samsung R\&D Institute India - Bangalore\\
Email: m.avinash@samsung.com}
\and
\IEEEauthorblockN{Anvesh Rao Vijjini} 
\IEEEauthorblockA{Samsung R\&D Institute India - Bangalore\\
Email: a.vijjini@samsung.com}}

%


\maketitle

\begin{abstract}
Elastic Weight Consolidation (EWC) is a technique used in overcoming catastrophic forgetting between successive tasks trained on a neural network. We use this phenomenon of information sharing between tasks for domain adaptation. Training data for tasks such as sentiment analysis (SA) may not be fairly represented across multiple domains. Domain Adaptation (DA) aims to build algorithms that leverage information from source domains to facilitate performance on an unseen target domain. We propose a model-independent framework - Sequential Domain Adaptation (SDA). SDA draws on EWC for training on successive source domains to move towards a general domain solution, thereby solving the problem of domain adaptation. We test SDA on convolutional, recurrent, and attention-based architectures. Our experiments show that the proposed framework enables simple architectures such as CNNs to outperform complex state-of-the-art models in domain adaptation of SA. In addition, we observe that the effectiveness of a harder first Anti-Curriculum ordering of source domains leads to maximum performance.\end{abstract}


%
\IEEEpeerreviewmaketitle

\section{Introduction}

\begin{figure}
\centering
  \includegraphics[width=4.5cm]{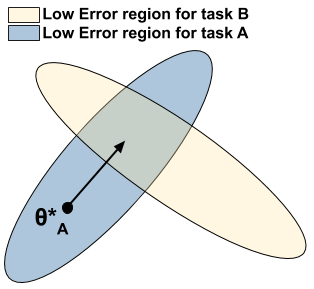}
  \caption{EWC.}
  \label{fig:ewc}
\end{figure}

In the continual learning framework of machine learning, a neural network performs poorly on tasks first trained on in a sequence of successive tasks. This problem is termed as catastrophic forgetting \cite{french1999catastrophic}. Recently, various approaches have been proposed to address it, such as Hard Attention (HT) \cite{serra2018overcoming} and Incremental Moment Matching (IMM) \cite{lee2017overcoming}. Elastic Weight Consolidation (EWC) \cite{kirkpatrick2017overcoming} is one such regularization based approach which has been shown to preserve performance over tasks effectively even after training on multiple tasks. Section \ref{sec:related_work} discusses other works in natural language processing (NLP) which have utilized EWC for overcoming catastrophic forgetting and more. However, in the proposed framework, we use EWC for a distinctive reason from catastrophic forgetting. EWC relies on the idea of over-parameterization\cite{kirkpatrick2017overcoming}. Over-parameterization suggests that a neural network has more than one optimal solution for the same task. Hence, given two tasks A and B, there can be a solution to task B, which exists in the low error space of task A. EWC tries to approach this solution for maximizing performance on both the tasks as seen in Figure \ref{fig:ewc}. The proposed framework Sequential Domain Adaptation (SDA), appropriates this idea for Domain Adaptation. We hypothesize that, for sentiment analysis on a specific domain, multiple solutions exist, at least one of which lies closest to the general domain optimal. We do not have any general domain data. Instead, we rely on continuous training across multiple domains to reach as close as possible to the general domain optimal, which gives the best performance on an unseen domain. Section \ref{sec:SDA and EWC} explains EWC and its utilization within the proposed framework.
We experiment SDA with varying continual learning strategies, ordering of domains for training, and across architectures to compare them against the state-of-the-art (SotA) models. Section \ref{sec:exp} details these experiments.
Our observations and experiments conclude that sequential training of domains in a harder-first approach or Anti-Curriculum domain ordering outperforms SotA. Curriculum Learning\cite{bengio2009curriculum} proposes that training a model such that it is provided easier samples first leads to better generalization. Anti-Curriculum, as the name suggests, advises that training with harder to easier examples leads to a better generalization. We discuss the effect Curriculum and Anti-Curriculum ordering along with other results in Section \ref{sec:results}.

The overall contributions of our paper are as follows:
\begin{itemize}
    \item We propose a framework SDA that outperforms SotA systems while employing a stricter resource-constrained definition of Domain Adaptation.
    \item Proposed framework is architecture invariant, enabling even simple and fast architectures to beat complex SotA architectures.
    \item Since the proposed framework draws from catastrophic forgetting, we compare catastrophic forgetting and domain adaptation by contrasting advanced continual learning methods with EWC, employed in SDA for Domain Adaptation.
    \item Observed results show the effectiveness of an Anti-Curriculum domain ordering for a better generalization across unseen domains.
\end{itemize}

\begin{figure}
\centering
  \includegraphics[width=4.5cm]{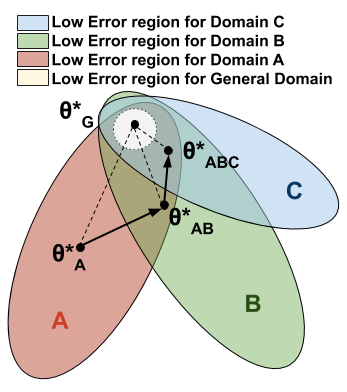}
  \caption{Proposed framework SDA that facilitates Domain Adaptation through EWC.}
  \label{fig:da}
\end{figure}

\section{Related Work}
\label{sec:related_work}
\textbf{Domain Adaptation:} Most existing domain adaptation works focus on learning general domain knowledge just from one domain rather than utilizing all. Recent SotA performance is recorded by pivot based methods \cite{ziser2018pivot,ziser2019task}, adversarial \cite{qu2019adversarial}, semi-supervised  \cite{he2018adaptive}, and domain classification \cite{liu2018learning} based techniques. We explain recent methods as baselines in Section \ref{sec:baselines} \\
\\\textbf{Elastic Weight Consolidation in NLP:}
In recent times EWC \cite{kirkpatrick2017overcoming} has been in used in various Natural Language Processing tasks, like in machine translation \cite{varis2019unsupervised,saunders2019domain,thompson2019overcoming}, Language Modeling \cite{wolf2018continuous} and Sentiment Analysis \cite{lv2019sentiment}.
Most of these works utilize Elastic Weight Consolidation to tackle catastrophic forgetting.
\cite{saunders2019domain,thompson2019overcoming} tackle Domain adaptation in Neural Machine Translation (NMT). They train their translation model on successive parallel corpora, with different domains and use EWC to retain performance on older domains. However, these works assume the existence of general domain data. \cite{wolf2018continuous} use EWC to reduce forgetting between distinct high level and low-level language modeling tasks. \cite{lv2019sentiment} leverage shared knowledge between successive domains with EWC to perform on trained domains the highest. They aim to improve sentiment analysis with the help of data from more domains. This idea has also been reflected in earlier works, such as \cite{yang2016leveraging}. However, the proposed work strictly keeps test domain unseen, as the aim is to use EWC to reach as close as possible to a general domain optimal solution with only seeing multiple training domain data. 

\begin{table}
\centering
\begin{tabular}{l|l|l|l|l} 
\hline
  & CNN & LSTM & A-LSTM & TE \\ 
\hline
\hline
$D$ &71.9 &70.95 &70.1 &64.25 \\ 
\hline
$E$ & 71.5& 61.2& 64.9&63.75  \\ 
\hline
$K$ & 64.9& 58.55&61.85 &58.95 \\ 
\hline
Init. & 60.5&63.5 &60.5 &59.75  \\ 
\hline
Comb & 73.4 &68.55 &65.85 &64.5  \\ 
\hline
IMM Mean &58.5 &52.7 &54.9 &57.35   \\ 
\hline
IMM Mode & 58.4& 54& 54.75&61.3 \\
\hline
HAT & 57.15& 55& 64.8&63.75  \\
\hline
\textbf{EWC} &72.35 &70.8 &\textbf{67.95} &\textbf{66.6}    \\
\hline
\end{tabular}
\caption{Comparison of SDA with various continual learning method and single training domain performance. In the approaches which entail a Domain Ordering (Init, IMM, HAT and EWC), best order results are reported. Test domain is Books.}
\label{tab:continual_books}
\end{table}

Furthermore, there have also been works on domain adaption, such as incremental domain adaptation (IDA) \cite{asghar2018progressive}. While our work resembles IDA in the idea of taking advantage of multiple domains for domain adaption, but it differs in the definition of domain adaptation. Within the IDA framework, the goal is to build a unified model that performs best on the domains observed so far by the model. Hence for \cite{asghar2018progressive}'s work overcoming catastrophic forgetting becomes an important task. However, in the proposed problem setting, the application of EWC is not to overcome catastrophic forgetting itself since we are not bothered with performance on source domains but rather to facilitate a framework to maximize return on an utterly unseen target domain. This difference between Domain Adaptation and Catastrophic Forgetting is discussed with results in Section \ref{sec:results_cataDA}
\section{Sequential Domain Adaptation Framework and Elastic Weight Consolidation}
\label{sec:SDA and EWC}
\subsection{Elastic Weight Consolidation}
The EWC \cite{kirkpatrick2017overcoming} loss function is devised to train on a task while not forgetting the parameters of a previous task. This is done by adding the difference of magnitude between current parameters and previous parameters weighted by the parameter's importance to the previous task.  Suppose a neural network is trained on task A, $\theta^*_{A}$  being the optimal value of the parameters on this task. Then, while training for task B, the following loss function will be used:
\begin{equation}
    \mathcal{L'}(\theta) = \mathcal{L}_B(\theta) + \sum_i \frac{\lambda}{2}F_i(\theta_i - \theta^*_{A,i})^2
\end{equation}
$\mathcal{L}_B(\theta)$ is the standard cross-entropy loss on task B in isolation. Within the EWC term itself. $\lambda$ is a hyperparameter. The summation $i$ is over all parameters with $\theta_i$ being a single parameter, and  $F_i$ is the diagonal element of corresponding Fisher information matrix of the parameter when training on task $A$. $F_i$ correlates with parameter gradients on task A squared $\left[ \frac{\partial \mathcal{L}_A(\theta_i)}{\partial \theta_i} \right]^2$. After training of task A, a higher gradient of a parameter implies that even small $\delta$ changes in the parameter's value will result to large changes in the predictions. Hence the model is sensitive to such parameters, reflecting their importance to task A. Conversely, lower gradient implies lower significance for task A. Hence, its regularization is weighted less and can be updated more flexibly when training task $B$.
\begin{table*}
\centering
\begin{tabular}{|l|l|l|l|l|} 
\hline
  & Kitchen (K) & Electronics (E) & DVD (D) & Books (B) \\ 
\hline
\hline
PBLM \cite{ziser2018pivot} &68.95 &59.4 &59.55 &61.4 \\ 
\hline
DSR \cite{liu2018learning} & 56&55 &56.3 & 52\\ 
\hline
BLSE \cite{barnes2018projecting}  & 80 & 74.25& \textbf{75.25}&71.25\\ 
\hline
DAS \cite{he2018adaptive} &70.75 &72.35 &60.85 &62.6 \\ 
\hline
ACAN \cite{qu2019adversarial}  &79.55 &73.65 &70.8 & \textbf{73.95} \\ 
\hline
\hline
\textbf{EWC-CNN} &\textbf{80.25} & 77.25&\textbf{73.8} & \textbf{72.35}\\
\hline
\textbf{EWC-LSTM} & 79.7& \textbf{78.25}& 71.35& 70.8\\
\hline
\textbf{EWC-ALSTM }& 75& 75.2& 69.05& 67.95\\
\hline
\textbf{EWC-TE} & 71.35&65.25 & 65.45& 66.6\\
\hline
\end{tabular}
\caption{Comparison of proposed framework SDA with state-of-the-art architectures on the Multi-Domain Sentiment Dataset. For the SotA, best performing source domain is chosen, while for SDA anti-curriculum source domain ordering is chosen.}
\label{tab:sota}
\end{table*}

Using Figures \ref{fig:ewc} and \ref{fig:da}, we explain our framework which enables us to exploit Elastic Weight Consolidation for Domain Adaptation completely invariant to the model itself. Figure \ref{fig:ewc} explains the original EWC reasoning. \cite{kirkpatrick2017overcoming} suggest that due to over parameterization, there exists a low error region in the parameter space for any task. All parameters within this space are equally optimal. When training on task B, using EWC we move to the common intersection with low error region of task B. We attune this idea in domain adaptation. Here each task could be interpreted as a domain. If there exists a general domain low error space, then successive EWC training across various domains will push the solution closest to this general domain low error space. Figure \ref{fig:da} demonstrates this idea. $\theta_{ABC}^{*}$ where A, B, and C are source domains model has been trained on, will be pushed close to $\theta_{G}^{*}$ where G is the ideal general domain for which we lack any actual data. As long as the training is done on enough number of domains we hypothesize that $||\theta_{ABC}^{*} - \theta_{G}^{*}||$ attenuates. Such a general domain solution will maximize performance on unseen domains solving the problem of domain adaptation. Since the framework only uses custom loss function from EWC, it is independent of the model architecture itself. 
\subsection{SDA} 
Given a set of training or ``source" domains $D_{j}$ where $j$ $\in$ $[1,n]$ and n is the number of source domains, the proposed framework is explained as follows. For a model architecture $f_{\theta}(x)$ and continual learning method $C$, f is trained iteratively on an ordered set of the source domains $<$$s$$>$. If the domain ordering strategy $<$$s$$>$ $ = (s_{1},s_{2},s_{3}....s_{n})$. where $s_{i}$ $\in$ $[1,n]$ then training is done sequentially in the order of $(D_{s_{1}},D_{s_{2}},D_{s_{3}}...D_{s_{n}})$. We test with all possible combinations of domains and report our results in Tables \ref{tab:cnn_lstm_order_kitchen} and \ref{tab:alstm_te_order_kitchen}. We get the best performance with $<$$s$$>$ being Anti-Curriculum or hardest first in nature. We explain the ordering strategy in Section \ref{sec:anticurr}. For every successive training between two domains $D_{t-1}$ and $D_{t}$, the parameter training of $f_{\theta}(x)$ is constrained by the continual learning method C such that performance of $f_{\theta}(x)$ on $D_{t-1}$ and $D_{t}$ is maximized.The aim of domain adaptation is to develop $f_{\theta}(x)$ to perform best on a completely unseen ``target" domain $D_{T}$ where $T \notin [1,n]$.  In SDA specifically, C is chosen to be EWC, but we experiment with other, more recent continual learning methods as well explained in Section \ref{sec:cata}. Algorithm \ref{algo} outlines SDA.

\begin{algorithm}
	\caption{Sequential Domain Adaptation} 
	\begin{algorithmic}[1]
	\Procedure{SDA}{$f_{\theta}$,$D_{j}$ where $j$ $\in$ $[1,n]$,$D_{T}$}
		\State Obtain $<$$s$$>$ $ = (s_{1},s_{2},s_{3}....s_{n})$
		\State Train $f_{\theta}$ on $D_{s_{1}}$ with Loss $\mathcal{L}_{D_{s_{1}}}$
		\State Obtain $\theta^*_{D_{s_{1}}}$ and $F$ from $f_{\theta}$
			\For{$t = (s_{2},s_{3}\ldots s_{n})$}
			   \State $\mathcal{EWC} = \sum_i \frac{\lambda}{2}F_i(\theta_i - \theta^*_{D_{t-1,i}})^2$
				\State Train $f_{\theta}$ on $D_{t}$ with $\mathcal{L}_{D_{t}} +\mathcal{EWC}$
				\State Obtain $\theta^*_{D_{t}}$ and $F$ from $f_{\theta}$
			\EndFor
			\State Test $f_{\theta}$ on $D_{T}$,  $T \notin [1,n]$ 
    \EndProcedure
	\end{algorithmic} 
\label{algo}
\end{algorithm}

\begin{table}
\centering
\begin{tabular}{l|l|l|l|l} 
\hline
  & CNN & LSTM & ALSTM & TE \\ 
\hline
\hline
$B$ & 71.25& 68.55 &67.55  &62.1 \\ 
\hline
$D$ & 71.08  &68.55  & 68 & 59.5 \\ 
\hline
$E$ & 79.05  & 78.3 & 73.25 & 70.8\\ 
\hline
Init. & 64.75 & 72.5 & 61.0 &58.25 \\ 
\hline
Comb & 73.95  & 73.5 & 69.05  & 67.2 \\ 
\hline
Mean & 58.45 &56.1 &53.7 &56.35  \\ 
\hline
Mode & 58.35& 56.85& 54.15&62.9 \\
\hline
HAT & 61.80&67.6 &67.5 &69.15 \\
\hline
\textbf{EWC} &\textbf{80.25} & \textbf{79.7} & \textbf{75.00} & \textbf{71.35}   \\
\hline
\end{tabular}
\caption{Comparison of SDA with various continual learning method and single training domain performance. In the approaches which entail a Domain Ordering (Init, IMM, HAT and EWC), best order results are reported. Test domain is Kitchen.}
\label{tab:continual_kitchen}
\end{table} 

\begin{table*}
\centering
\begin{tabular}{|l|llllll|llllll|} 
\hline
\multirow{2}{*}{$C$} & \multicolumn{6}{c|}{CNN} & \multicolumn{6}{c|}{LSTM} \\ 
 & BDE & BED & \textbf{DBE} & DEB & EBD & EDB & BDE & BED & \textbf{DBE} & DEB & EBD & EDB \\ 
\hline
\hline
Init & 64.75 &62.75 &64.25 &59 &59.25 & 51.75 & 72.5 & 59&66.5 &61.25 &62.5 &57.50  \\
Mean & 56&57.6 & 58.45 &53.15 &55.6 &54.15 &56.1 &54.5 &55.9 &54.55 &52.9 &52.1  \\
Mode &58.35 &51 &55.1 & 54.7 & 54.6 & 54.4 & 55.05& 55.25&56.85 &53.35 &53 &52.25 \\
HAT & 61.80&56.55 &61.15 &60.30 &58.35 &53.45 &66.2 &53.5 &67.6 &51.40 &51.50 &53.2  \\
\textbf{EWC} &79.4 &72.1 &\textbf{80.25} &69.20 &71.6 &69.20  & \textbf{79.7} &68.5 & \textbf{79.7} &68.5 &68.5 &68.5  \\
\hline
\end{tabular}
\caption{Effect of domain ordering across various continual learning methods. Target domain is Kitchen.}
\label{tab:cnn_lstm_order_kitchen}
\end{table*} 

\begin{table*}
\centering
\begin{tabular}{|l|llllll|llllll|} 
\hline
\multirow{2}{*}{$C$} & \multicolumn{6}{c|}{ALSTM} & \multicolumn{6}{c|}{TE} \\ 
 & BDE & BED & \textbf{DBE} & DEB & EBD & EDB & BDE & BED & \textbf{DBE} & DEB & EBD & EDB \\ 
\hline
\hline
Init &61&55.25 &59.5 &49 &57.50 &54.75  &58 &55.75 &58.25 &46.50 &54.25 &54.50 \\
Mean & 52.75&52.15 &53.7 &51.05 &52.35 &51.85 & 54.35& 56.35&55.25 &51.8 &54.65 &52.2  \\
Mode &54.15 &51.6 &54.1 &51.85 &52.75 &52.1 &62.9 &55.9 &59.05 &58.7 &56.05 &56.55\\
HAT & 66.35& 60.35& 67.50& 58.90&61.20 &59.95 & 69.15& 61.35&67.55 &61.45 &62.40 &62.75\\
\textbf{EWC} & \textbf{75} & 65.40 & \textbf{75} &66.65 &65.4 &66.65 &\textbf{71.35} &59.60 & \textbf{71.35} &61.85 &59.60 &61.85  \\
\hline
\end{tabular}
\caption{Effect of domain ordering across various continual learning methods. Target domain is Kitchen.}
\label{tab:alstm_te_order_kitchen}
\end{table*}

In the following sections, we explain our experimental details and choices of the model architectures $f$, the continual learning method $C$, and the domain ordering $<$$s$$>$. 
\section{Experimental Setup}
\label{sec:exp}
\subsection{Datasets}
We perform experiments on the standard Multi-Domain Sentiment Dataset \cite{blitzer2007biographies}. It contains reviews from 4 domains, namely Books (B), DVD (D), Electronics (E), and Kitchen (K). Each domain has 2000 reviews, of which 1000 reviews belong to the positive polarity, and 1000 reviews belong to negative polarity. In each domain, 1280 reviews are used for training, 320 for validation, and 400 for testing. All the reported results are averaged over five runs. While reporting results on a target domain, the training sequence of domains $D_{1}$, $D_{2}$ and $D_{3}$ is represented as $D_{1}D_{2}D_{3}$.
\subsection{Architectures}
\label{sec:arch}
We experimented the following neural network architectures $f$ for performing SDA. Hyperparameters of the Architectures are specified in Section \ref{sec:implementation}:
\subsubsection{CNN}
This architecture is based on the popular CNN-non-static model used for sentence classification \cite{kim2014convolutional}. 
\subsubsection{LSTM}
We used a single layer LSTM model \cite{hochreiter1997long}, and the output at the last time-step is fed into a fully connected output layer.
\subsubsection{Attention LSTM (ALSTM)}
This architecture is same as the above LSTM, except the attention \cite{bahdanau2014neural} mechanism is applied on the outputs across all time-steps. The weighted sum is fed into a fully connected output layer.
\subsubsection{Transformer Encoder (TE)}
This architecture uses a Bidirectional Transformer also known as Transformer Encoder \cite{vaswani2017attention}. Transformer encoder has been very popularly used for text classification \cite{devlin2018bert}. The output from the Encoder is averaged across time-steps and fed into a fully connected output layer. We do not initialize our architecture with the weights of popular transformer encoders such as BERT\cite{devlin2018bert}. These models first pretrain their architecture on large general domain data. However, proposed models assume the absence any such general data. Hence, for a fairer comparison the transformer encoder is randomly initialized.

\subsection{Continual Learning Baselines}
\label{sec:cata}
We compare our approach with the following continual learning methods $C$. Note that these baselines adapt the same architectures $f$ mentioned in section \ref{sec:arch}. The training procedure of SDA differs among these baselines.
\subsubsection{Weight Initialization (Init)}
Let $D_{i-1}$, $D_{i}$ be the domains trained sequentially using the neural network architecture $f$. The parameters of $f$ for training on $D_{i}$ are initialized with the parameters of $f$ trained on $D_{i-1}$.
\subsubsection{Combined Training (Comb)}
In this baseline, the data from the three domains are combined to train the neural network $f$ just once. 
\subsubsection{Incremental Moment Matching (IMM)}
Incremental Moment Matching (IMM) \cite{lee2017overcoming} is a method proposed for overcoming catastrophic forgetting between tasks. It incrementally learns the distribution of neural network trained on subsequent tasks. We compare our approach with two IMM techniques IMM-Mean (Mean) and IMM-Mode (Mode). 
\subsubsection{Hard Attention to Task (HAT)}
Hard Attention to Task (HAT) \cite{serra2018overcoming} is also proposed for overcoming catastrophic forgetting between tasks. A hard attention mask is learned concurrently to every task, which preserves the previous task's information without affecting the learning of current task.
\subsection{State-of-the-art Baselines}
\label{sec:baselines}
The URLs of the code used for performing experiments on SotA architectures are given in the Appendix
\subsubsection{PBLM}
Pivot Based Language Model (PBLM) \cite{ziser2018pivot} is a representation learning model that combines pivot-based learning with Neural Networks in a structure-aware manner. The output from PBLM model consists of a context-dependent representation vector for every input word.
\subsubsection{DSR}
\cite{liu2018learning} learns Domain-Specific Representations (DSR) for each domain, which are then used to map adversarial trained general Bi-LSTM representations into domain-specific representations. This domain knowledge is further extended by training a memory network on a series of source domains. This memory network holds domain-specific representations for each of the source domains.
\subsubsection{BLSE}
Bilingual Sentiment Embeddings (BLSE) \cite{barnes2018projecting} casts Domain Adaptation problem as an embedding projection task. The model takes input as embeddings from two domains and projects them into a space representing both of them. This projection is jointly learned to predict the sentiment.

\begin{table*}
\centering
\begin{tabular}{|l|llllll|llllll|} 
\hline
\multirow{2}{*}{$C$} & \multicolumn{6}{c|}{CNN} & \multicolumn{6}{c|}{LSTM} \\ 
 & \textbf{DBK} & DKB & BDK & BKD & KDB & KBD & \textbf{DBK} & DKB & BDK & BKD & KDB & KBD \\ 
\hline
\hline
Init & 70&60 &72.25 &60.5 &63.75 &62.25 & 73.75 &61 &69.25 &64.5 &59.75 & 59.75\\
Mean & 59.15&54.45 &60 &55.6 &55.25 &56 & 55.7 &50.95 &55 &53.65 &52.45 &52.4 \\
Mode & 58.85&54.4 &56.3 &53.3 &54.8 &53.8 & 52.9 &52.95 &55.7 &51.75 &52.05 &54 \\
HAT &66.1 &52.2 & 67.2& 60& 54.85&58.55   &62.9   &55.05 &62.3 &59.2 &54.15 & 58.45\\
\textbf{EWC} & \textbf{77.25}&71.55 &76.95 &71.8 &71.55 &71.6 & \textbf{78.25}  &68.05 & \textbf{78.25} &67.7 &68.05 & 67.7\\
\hline
\end{tabular}
\caption{Effect of domain ordering across various continual learning methods. Target domain is Electronics.}
\label{tab:cnn_lstm_order_electronics}
\end{table*}

\begin{table*}
\centering
\begin{tabular}{|l|llllll|llllll|} 
\hline
\multirow{2}{*}{$C$} & \multicolumn{6}{c|}{ALSTM} & \multicolumn{6}{c|}{TE} \\ 
 & \textbf{DBK} & DKB & BDK & BKD & KDB & KBD & \textbf{DBK} & DKB & BDK & BKD & KDB & KBD \\
\hline
\hline
Init & 65&55.75 &70.5&58.25&58.5 &59.75 &57.5  &59.25 &53 &52.75 &52&60 \\
Mean &54.35 &53.35 &53.9&52.2&50.95 &52.3 & 60.1 &53.35 &58.45 &55.55 &54.2&56.55 \\
Mode & 52.45&51.3 &53.7&51.2&51.6 &51.35 & 61.65 &57.9 &63.1 &54.15 &52.35& 55.4\\
HAT &72 &64 &72.75 &62.05 &64.05 & 63.15  & 65.25  &59.2 &65.25 &60.65 &59.2 &60.65 \\
\textbf{EWC} & \textbf{75.2}&66.45 &\textbf{75.2} &69.35 &66.45 &69.35 &\textbf{72.55 } &61.6 &\textbf{72.6} &64.85 &62.5 & 63.55\\
\hline
\end{tabular}
\caption{Effect of domain ordering across various continual learning methods. Target domain is Electronics.}
\label{tab:alstm_te_order_electronics}
\end{table*}

\subsubsection{DAS}
Domain Adaptive Semi-supervised learning (DAS) \cite{he2018adaptive} minimizes the distance between the source and target domains in an embedded feature space. For exploiting additional information about target domain, unlabelled target domain data is trained using semi-supervised learning. This is done by employing the regularization methods of entropy minimization and self-ensemble bootstrapping.

\subsubsection{ACAN}
Adversarial Category Alignment Network (ACAN) \cite{qu2019adversarial} trains an adversarial network using labeled source data and unlabelled target data. It first produces ambiguous features near the decision boundary reducing the domain discrepancy. A feature encoder is further trained to generate features appearing at the decision boundary.


\begin{figure*}
\centering
  \includegraphics[width=14cm]{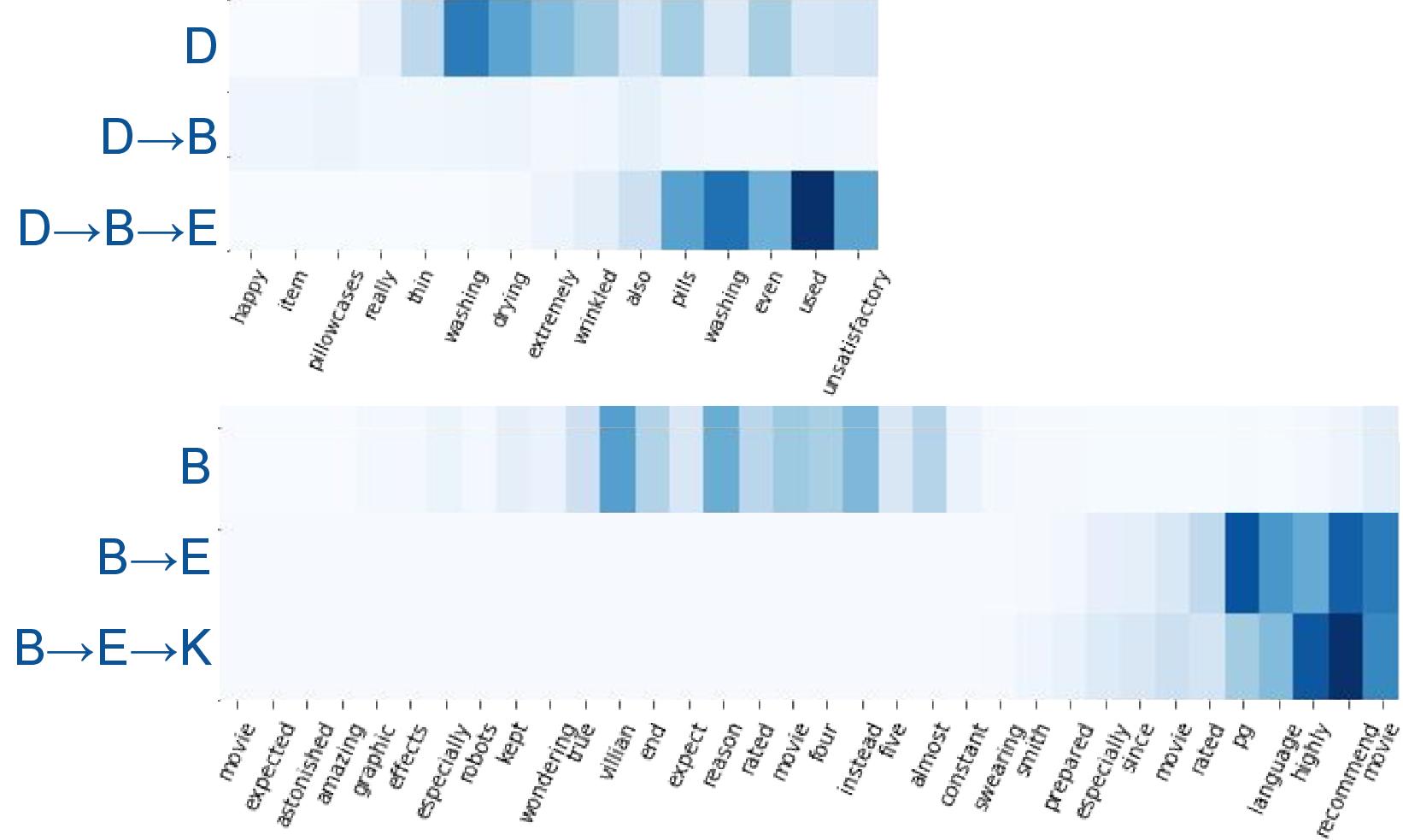}
  \caption{Attention visualizations from ALSTM model.}
  \label{fig:aten}
\end{figure*}

\section{Implementation Details}
\label{sec:implementation}

Let $I$ denote the input sentence represented by a sequence of words $I = \{ i_{1},i_{2}.i_{3},....,i_{n}\} $ where $n$ is the maximum sentence length. Let $V$ be the vocabulary of words and $X \in R^{n \times d}$ denote the pretrained word embeddings where $d$ denotes the dimensions of word embeddings. Words present in the vocabulary $V$ are initialized to the corresponding word embeddings and words not present are initialized to 0's and are updated during training. Therefore, input $I$ is converted to $W \in R^{B \times d}$. We used a maximum sentence length of 40 and glove 840B pretrained embeddings\footnote{https://nlp.stanford.edu/projects/glove/}.
\subsection{CNN Architecture}
Convolutional layers with kernel sizes 3, 4 and 5 are applied on the input $X$ simultaneously. The number of filters are 100 and activation function in the convolutional layer is ReLU. The outputs from each of the convolutional layer are concatenated and fully connected to the sigmoid layer. A dropout of 0.5 is applied on the fully connected layer. CNN is trained for 30 epochs with a batch size of 16. Early-stopping mechanism is applied if the validation loss doesn't decrease for 10 epochs. AdamW is used as the optimizer with a learning rate of 0.001.
\subsection{LSTM Architecture}
The input $X$ is passed through a single layer LSTM. Hidden dimensions of LSTM is 100. LSTM is trained for 25 epochs with a batch size of 15. AdamW is used as the optimizer with a learning rate of 0.001.
\subsection{ALSTM Architecture}
The input $X$ is passed through the single layer LSTM. Attention mechanism is applied on the outputs of all the timesteps. Hidden dimensions of LSTM are 100 and hidden layer dimension of attention mechanism is 64. The output from attention is fully connected to the sigmoid output layer. ALSTM is trained for 30 epochs with a batch size of 35. AdamW is used as the optimizer with a learning rate of 0.0081.
\subsection{Bi-directional Transformer Encoder Architecture (TE)}
The input $X$ is passed through a Bi-directional Transformer Encoder. The outputs from the transformer encoder are averaged across timesteps and are fully connected to sigmoid output layer. Number of Encoder layers are 2, number of heads in the multihead attention models are 5, the dimension of the feedforward network model is 256. TE is trained for 30 epochs with a batch size of 35. AdamW is used as the optimizer with a learning rate of 0.001.

\section{Results and Discussion}
\label{sec:results}
Figure \ref{fig:aten} visualizes the attention scores from the ALSTM model of samples from Kitchen (K) and DVD (D) target domains. The attention scores are shown at each step in SDA when the model encounters a new source domain. In these examples, the model correctly classifies the sentiment on only observing the third source domain. As explained in Section \ref{sec:SDA and EWC}, the target domain is always unseen by the model. These examples help us understand how SDA works internally. We note the following crucial points.
\begin{itemize}
    \item In the first example where Kitchen is the target domain, on just observing DVD, the model attends to ``thin", ``washing" and ``drying". These words describe pillowcases, even though the sentence is a washing machine review, attending on them produces an incorrect positive outcome. On encountering the second domain, the model learns that the actual sentiment of the sentence is not conveyed in these terms. However, it is only on supervising the Electronics domain, the attention is focused on the crucial terms ``used" and ``unsatisfactory", giving the real sentiment.\footnote{As part of preprocessing, stop-words have been removed.}

    \item In the second example where DVD is the target domain, on encountering just Books, the focus is arbitrary and hence resulting in the wrong prediction. The focus is just right on encountering Electronics, and precisely on the ``highly recommended movie" when encountering the final source domain.
    \item In both examples, we see how the model with sequential EWC training, moves from arbitrary domain-specific attention to learning more general domain knowledge, helping it predict a sample of an unseen domain. Without the framework, the model fails to generalize from a single source domain and ends up making incorrect predictions.
\end{itemize}
In our experiments described in Section \ref{sec:exp}, proposed approach comparison with the SotA architectures is discussed, the results of these experiments have been charted in Table \ref{tab:sota}. The choice of continual learning method EWC has been justified in Tables \ref{tab:continual_books} ,\ref{tab:continual_kitchen} ,\ref{tab:continual_electronics} and \ref{tab:continual_dvd}. Furthermore, Tables \ref{tab:cnn_lstm_order_kitchen}, \ref{tab:alstm_te_order_kitchen} and \ref{tab:cnn_lstm_order_electronics}, \ref{tab:alstm_te_order_electronics} compare the anti-curriculum order of training with other orders across models in SDA.
\subsection{SDA and State-of-the-Art Domain Adaptation}
As shown in Tables \ref{tab:sota} and , the models trained with SDA framework outperform the recent state-of-the-art in Domain Adaptation of sentiment Analysis. PBLM, DAS, ACAN \cite{ziser2019task,ziser2018pivot,he2018adaptive,qu2019adversarial} use Domain adaptation in a semi-supervised setting. In other words, they use large quantities of target domain unlabelled data for training their respective architecture. Our framework strictly keeps the target domain unseen and still can outperform them except for ACAN in one target domain. This shows the effectiveness of SDA framework in resource scarce setting. DSR, similar to the proposed framework, uses multiple source domains for learning domain representations. Their model relies heavily on domain classification. However, training a robust domain classifier requires much more data, hence in our resource-constrained setting, they give a poor performance. While BLSE outperforms SDA in one target domain, they utilize labelled target domain data for training their architecture, whereas as SDA keeps target domain strictly unseen. This makes the domain adaptation definition of proposed framework much more stringent. Apart from DSR, all other architectures observe target domain in some way. The EWC regularized training of SDA ameliorates the requirement to observe the target domain. EWC-CNN and EWC-LSTM closely follow the highest results when not the highest themselves. Other architectures, EWC-ALSTM and EWC-TE might be poor because of small dataset size, failing to generalize on unseen domains.


\subsection{Effectiveness of Anti-Curriculum order of Training}
\label{sec:anticurr}
Since SDA facilitates training in a sequential manner, there is a multitude of orders in which the model can observe the domains. Specifically, $n$ domains entail $n!$ orders. Tables \ref{tab:cnn_lstm_order_kitchen}, \ref{tab:alstm_te_order_kitchen} and \ref{tab:cnn_lstm_order_electronics}, \ref{tab:alstm_te_order_electronics} show the performance on all possible domain orderings kitchen and electronics as test domains respectively\footnote{Due to constraint in page length, we have kept the domain ordering tables for DVD and Books test domain in appendix here: \href{https://drive.google.com/file/d/1RX_5zNtPkJgQaow-I5MtFZTqReWhExNb/view?usp=sharing}{Here}.}. For EWC, we observe that a certain ordering has largely performed best. On comparing the single source domain setting in Table \ref{tab:continual_kitchen}, we see that for Kitchen as a target domain, Electronics is the easiest followed by Books and DVD. Hence, DBE can be characterized as ``Anti-Curriculum" strategy where the hardest examples are provided first followed by easier ones. Conversely, EBD becomes a curriculum strategy. Followed closely by DBE are the results of BDE, this shows that while anti-curriculum works best, there is more bias towards the last domain and not all positions in the ordering are equally important. 
While the effect of curriculum strategies in text classification is well documented \cite{cirik2016visualizing,han2017tree}, the result of anti-curriculum strategies in literature has been mixed. \cite{hacohen2019power,weinshall2018curriculum} show that anti-curriculum strategies work worst among no curriculum and curriculum. However, anti-curriculum effectiveness has been demonstrated by some, such as \cite{mccann2018natural}. Our results demonstrate that an anti-curriculum ordering of domains works best and furthermore, curriculum ordering gives one of the most unsatisfactory outcomes, even less than results when the model observes a single domain as shown in Table \ref{tab:sota}. This indicates that the choice of curriculum or anti-curriculum is heavily task-dependent. Previous\cite{hacohen2019power,weinshall2018curriculum,mccann2018natural} and current work show that if either of curriculum or anti-curriculum work, the converse strategy leads to a reduced performance than no curriculum.

\begin{table}
\centering
\begin{tabular}{l|l|l|l|l} 
\hline
  & CNN & LSTM & A-LSTM & TE \\ 
\hline
\hline
$B$ &70.55 &66.7 &69.95 &58.55 \\ 
\hline
$D$ & 72.3& 77&63.9 &58.55  \\ 
\hline
$K$ &77.1 &77.65 &72.95 &64.3 \\ 
\hline
Init. & 72.25& 73.75&70.5 &60  \\ 
\hline
Comb & 78&75.45 &75.6 &68.3  \\ 
\hline
IMM Mean & 60&55.7 &54.35 &60.1   \\ 
\hline
IMM Mode &58.85 &55.7 &53.7 &63.1 \\
\hline
HAT & 67.2& 62.9&72.75 &72.6  \\
\hline
\textbf{EWC} & 77.25&\textbf{78.25} &75.2 &65.25    \\
\hline
\end{tabular}
\caption{Comparison of SDA with various continual learning method and single training domain performance. In the approaches which entail a Domain Ordering (Init, IMM, HAT and EWC), best order results are reported. Test domain is Electronics.}
\label{tab:continual_electronics}
\end{table}

\begin{table}
\centering
\begin{tabular}{l|l|l|l|l} 
\hline
  & CNN & LSTM & A-LSTM & TE \\ 
\hline
\hline
$B$ & 73.8 & 70.55&66.8 &63.6 \\ 
\hline
$E$ & 68.7& 64&65 &61.85  \\ 
\hline
$K$ &69.25 &65.1 &65 &62.35 \\ 
\hline
Init. & 58.75&59.25 &62.25 &59  \\ 
\hline
Comb &71.9 &68.75 &67.1 &62  \\ 
\hline
IMM Mean & 58.5&52.75 &54.85 &55.6   \\ 
\hline
IMM Mode & 56.6&53.45 &52.05 &62.95 \\
\hline
HAT &59.2 &56.85 &62.8 &63.8  \\
\hline
\textbf{EWC} & \textbf{73.8} & \textbf{71.35} & \textbf{69.05} &\textbf{65.45}    \\
\hline
\end{tabular}
\caption{Comparison of SDA with various continual learning method and single training domain performance. In the approaches which entail a Domain Ordering (Init, IMM, HAT and EWC), best order results are reported. Test domain is DVD.}
\label{tab:continual_dvd}
\end{table}

\subsection{Catastrophic Forgetting and SDA}
\label{sec:results_cataDA}
The problem of Catastrophic Forgetting and Domain Adaptation within the SDA framework are fundamentally different. This is shown in Tables \ref{tab:continual_books}, \ref{tab:continual_kitchen}, \ref{tab:continual_electronics} and \ref{tab:continual_dvd} where various recent continual learning methods are compared against EWC. Previous works \cite{serra2018overcoming,lee2017overcoming} have established that these methods outperform EWC at overcoming catastrophic forgetting and are better at remembering multiple tasks learnt in a continual learning setting. However, across various architectures, we find that these methods are not good at Domain Adaptation. Despite being able to remember previous domain information, they fail to push model parameters closer to a general domain low error region, solving the problem of domain adaptation. These methods perform even weaker than settings where the model only encounters a single source domain. A Domain Adaptation setting much more dependent to catastrophic forgetting is IDA\cite{asghar2018progressive} where the model can only train on a single domain at a time, similar to proposed SDA, however, the model needs to perform best on all trained domains. Hence remembering disparate domain knowledge observed by the model becomes key to building solutions here. 
One question that arises when we study the SDA framework is that \textit{"If the model is observing multiple source domain data anyway, then why don't we combine all $D_{G}$ = $\{D_{1},D_{2} \ldots D{n}\}$ and call $D_{G}$ as General Domain data and train on it.``}. However, as we see in Table \ref{tab:continual_kitchen} proposed framework outperforms combined baseline across all architectures. This is because, as long as target domain $D_{T}$ $\notin$ $D_{G}$, $D_{G}$ is never really the General Domain. A model $f$, trained on $D_{G}$ may perform sound on all the domains contained within $D_{G}$ but not outside it. As explained in Figures \ref{fig:ewc} and \ref{fig:da}, in SDA, using EWC we move to a solution space as close as possible to the actual general domain solution for which we can never have the data since target domain is unseen.
\begin{figure}
\centering
  \includegraphics[width=8cm]{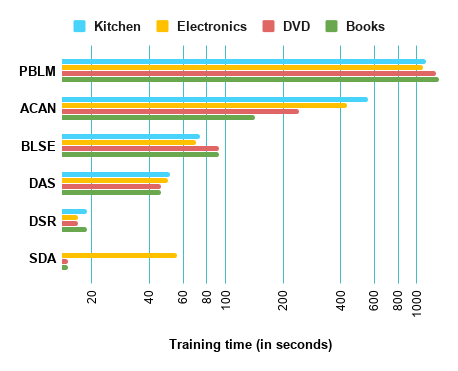}
  \caption{Training time comparison. Presented times for from results shown in Table \ref{tab:sota}. Please note that this graph has been visualized in the logarithmic scale.  }
  \label{fig:time}
\end{figure}

\subsection{Training time of SotA models}
Since the proposed framework enables even simple and low parameter models such as CNN to outperform state-of-the-art models, we also get an advantage in training time. This is shown in Figure \ref{fig:time}. Apart from DSR, all architectures use only a single target domain and yet take much more time to train. Among proposed architectures for SDA, CNN not only performs best as we saw in Table \ref{tab:sota}, but also is the quickest to train by a large margin. Comparatively on Electronics domain, we see LSTM takes five times more time than CNN. However, it is still five to thirty times less than PBLM and ACAN. As we saw in Table \ref{tab:sota}, BLSE and ACAN outperformed SDA in DVD and Books. However, they are also ten times slower than EWC-CNN, which take 15 and 16 seconds respectively. This shows the efficiency of SDA at empowering low parameter models such as CNN to match large architectures.

\section{Conclusion}In this paper, we present a new framework SDA to enable Domain Adaptation for Sentiment Analysis models. The proposed framework utilizes data from multiple domains but strictly keeps target domain data unseen. The approach aims models to be trained sequentially on the source domains with Elastic Weight Consolidation between successive training steps. In doing so, the framework empowers even simple model architectures to outperform complex state-of-the-art systems. SDA is tested on various models showing its independence on architectures. An anti-curriculum ordering of domains leads to the best performance. However, the shortcoming of such an ordering is that it necessitates the requirement of all source domains beforehand. Including that they were individually tested against the target domain to compute their respective difficulty or easiness. Future work could include how to make SDA invariant to the domain ordering as well.






\bibliographystyle{IEEEtran}
\bibliography{IEEEabrv.bib}
%

\section{Appendix: Results and Discussions}
\label{sec:experiments}
The effect of domain ordering in these algorithms are presented for DVD target domain in Tables \ref{tab:cnn_lstm_order_dvd} and \ref{tab:alstm_te_order_dvd} and for Books target domain in Tables \ref{tab:cnn_lstm_order_books} and \ref{tab:alstm_te_order_books} across all architectures. As we can observe from the aforementioned Tables, our hypothesis from the results on Kitchen and Electronics Target domains stand true across other testing domains as well.

\begin{table*}
\centering
\fontsize{10}{12}\selectfont
\begin{tabular}{|l|llllll|llllll|} 
\hline
\multirow{2}{*}{$C$} & \multicolumn{6}{c|}{CNN} & \multicolumn{6}{c|}{LSTM} \\ 
 & BEK & BKE & EBK & \textbf{EKB} & KBE & KEB & BEK & BKE & EBK & \textbf{EKB} & KBE & KEB \\ 
\hline
\hline
Init & 57.5& 56.75&58.75&57.25 &57.25&56 &53.5 &58  &54.25 &50.25 &59.25 &53\\
Mean & 54.9&51.6 &56.65 &58.5 &53.5 &56.7 &52.4  &50.9 &52.3 &52.75 &51.9&51.3 \\
Mode & 53.5&53.4 &52&51.7&54.3 &56.6 &52.95  &52.5 &53.45 &53.25 &52.8&52.65 \\
HAT & 58.35&53.95 &56.9 &58.85 &55.35 &59.2   &54.7   &55.5 & 54.05&55.9 &54.7 &56.85 \\
\textbf{EWC} & 70.55& 69.8& 70.85&\textbf{73.8} &69.95 &73.15   &66.7   &63.25 & 66.7&\textbf{71.35} &63.25 &\textbf{71.35} \\
\hline
\end{tabular}
\caption{Effect of domain ordering across various continual learning methods. Target domain is DVD.}
\label{tab:cnn_lstm_order_dvd}
\end{table*} 
\begin{table*}
\centering
\fontsize{10}{12}\selectfont
\begin{tabular}{|l|llllll|llllll|} 
\hline
\multirow{2}{*}{$C$} & \multicolumn{6}{c|}{ALSTM} & \multicolumn{6}{c|}{TE} \\ 
 & BEK & BKE & EBK & \textbf{EKB} & KBE & KEB & BEK & BKE & EBK & \textbf{EKB} & KBE & KEB \\
\hline
\hline
Init & 62.25&58.5 &56.75&57.25& 53.5& 61.75& 56 &55.25 &52 &55.5 &59&52.5 \\
Mean & 25.5&52.2 &52.8&53.7&54.85 &53.4 & 55.6 &53.05 &54.65 &53.4 &54.1&54.3 \\
Mode &50.85 &52 &50.8&51.3&51.55 &52.05 & 57.7 &59.45 &56.05 &59 &55.3&62.95 \\
HAT & 59&62.5 &59.8 &62.45 &62.8 &62.25   & 59.5  &61.85 &60.35 &62.95 &62.7 &63.8 \\
\textbf{EWC} & 65.5&64.85 &65.5 &\textbf{69.05} &64.85 &\textbf{69.05}   &  60.1 &60.7 &60.1 &\textbf{65.45} &60.7 &\textbf{65.45} \\
\hline
\end{tabular}
\caption{Effect of domain ordering across various continual learning methods. Target domain is DVD.}
\label{tab:alstm_te_order_dvd}
\end{table*}

\begin{table*}
\centering
\fontsize{10}{12}\selectfont
\begin{tabular}{|l|llllll|llllll|} 
\hline
\multirow{2}{*}{$C$} & \multicolumn{6}{c|}{CNN} & \multicolumn{6}{c|}{LSTM} \\ 
 & DEK & DKE & EDK & \textbf{EKD} & KDE & KED & DEK & DKE & EDK & \textbf{EKD} & KDE & KED\\ 
\hline
\hline
Init &57 &57.75 &56.75&60.5&54.75 &60.5 &61  &56.5 &61.5 &58.5 &56.25&63.5 \\
Mean & 54.5&54.1 &58.5&56.95&53.2 &57.2 &52.1  & 49.6& 52.7&51.2 &51.15&51 \\
Mode &55.25 &57.4 &53.2&53.4&57.45 &58.4 &52.9  &53.3 &54 &53.1 &53.05&52.7 \\
HAT &54.6 &54.4 &56.45 &57.15 &52.6 &57   &54.65   &55.15 &54.8 &55.05 &55.7 &54.65 \\
\textbf{EWC} &66.35 &69.75 &66.25 &\textbf{72.35} &70.2 &72.1   &58.65   & 61.75&58.65 &\textbf{70.8} &61.75 &\textbf{70.8} \\
\hline
\end{tabular}
\caption{Effect of domain ordering across various continual learning methods. Target domain is Books.}
\label{tab:cnn_lstm_order_books}
\end{table*} 
\begin{table*}
\centering
\fontsize{10}{12}\selectfont
\begin{tabular}{|l|llllll|llllll|} 
\hline
\multirow{2}{*}{$C$} & \multicolumn{6}{c|}{ALSTM} & \multicolumn{6}{c|}{TE} \\ 
 & DEK & DKE & EDK & \textbf{EKD} & KDE & KED & DEK & DKE & EDK & \textbf{EKD} & KDE & KED\\  
\hline
\hline
Init & 53.25&51.5 &56.25&60&56.5 &60.5 & 52.5 &59.75 &54 &54.25 &50.25&53 \\
Mean & 53.45&53.55 &52.65&53.35& 54.9&51.15 & 55.85 &55.15 &57.35 &56.6 &55.75&55.2 \\
Mode & 53.3&53.45 &53.8&54.75&53.15 &54.15 & 61.3 &59.3 &56.5 &55.65 &60.85&57.1 \\
HAT & 60.35&59.65 &60.5 &64.55 &59.75 &64.8   &  62.05 &60.05 &61.35 &63.45 &59.7 &63.75 \\
\textbf{EWC} & 60.9&65.75 &60.9 &\textbf{67.95} &65.75 & \textbf{67.95}  & 56.7  &62.25 & 56.7& \textbf{66.6}&62.25 &\textbf{66.6} \\
\hline
\end{tabular}
\caption{Effect of domain ordering across various continual learning methods. Target domain is Books.}
\label{tab:alstm_te_order_books}
\end{table*}
\subsection{Baseline Architectures}
We provide URLs of the code used for running Baseline Architectures and Continual Learning methods.
\\ \textbf{IMM Mean and Mode} 
\\https://github.com/joansj/hat/tree/master/src/
approaches
\\ \textbf{HAT}\\
https://github.com/joansj/hat/tree/master/src/
approaches
\\ \textbf{PBLM}\\
https://github.com/yftah89/PBLM-Domain-Adaptation
\\ \textbf{DSR}\\
https://github.com/leuchine/multi-domain-sentiment
\\ \textbf{BLSE}\\
https://github.com/jbarnesspain/domain\_blse
\\ \textbf{DAS}\\
https://github.com/ruidan/DAS
\\ \textbf{ACAN} \\
https://github.com/XiaoYee/ACAN
\\ \textbf{TRL-PBLM} \\
https://github.com/yftah89/TRL-PBLM

\subsection{SotA Implementations}
We used a maximum sentence length of 40 for DAS and ACAN instead of 3000. We fine-tuned PBLM by training for 30 epochs and early stopping is imposed with a patience of 10. IMM Mean, IMM Mode and HAT are trained with a batchsize of 32 instead of 64

\end{document}